%% file: main.tex
\ifdefined\pdfminorversion\pdfminorversion=7\fi
\documentclass[11pt]{article}
\usepackage[preprint]{acl}
\usepackage{times}
\usepackage{latexsym}
\usepackage[T1]{fontenc}
\usepackage[utf8]{inputenc}
\usepackage[protrusion=false,expansion=false]{microtype}
\usepackage{graphicx}
\usepackage{booktabs}
\usepackage{xcolor}

\usepackage{colortbl}
\definecolor{lengthband}{RGB}{230,230,255}
\usepackage{amsmath}
\usepackage{amssymb}
\usepackage{placeins}
\usepackage{tikz}
\usetikzlibrary{calc,arrows.meta}
\definecolor{extrapOrange}{RGB}{196,110,30}
\newcommand{\extrapStrong}{\textcolor[RGB]{35,125,65}{\ensuremath{\checkmark}}}
\newcommand{\extrapPartial}{\textcolor{extrapOrange!85!black}{\ensuremath{\sim}}}
\newcommand{\extrapLimited}{\textcolor[RGB]{185,45,40}{\ensuremath{\times}}}

\title{Explaining Length Extrapolation Across Attention Architectures}
\author{
Xinying Hu\textsuperscript{1,2},
Xiang Hu\thanks{Corresponding author.}\textsuperscript{3},
Xinyu Wei\textsuperscript{1,2},
Kewei Tu\footnotemark[1]\textsuperscript{1,2} \\
\textsuperscript{1}School of Information Science and Technology, ShanghaiTech University \\
\textsuperscript{2}Shanghai Engineering Research Center of Intelligent Vision and Imaging \\\textsuperscript{3}Tencent \\
\texttt{\{huxy2023, weixy2024, tukw\}@shanghaitech.edu.cn} \\
\texttt{shawnxxxhu@tencent.com}
}

\begin{document}
\raggedbottom
\maketitle

\begin{abstract}

Long-context modeling supports document understanding and long-horizon interaction. Selective access to the key--value cache offers one way to process very long contexts, with recent methods enabling retrieval far beyond the training window. However, poor length generalization in attention limits the context length over which models can reliably retrieve information. Although prior work has improved extrapolation through positional encoding, attention scaling, and sparse retrieval, some methods remain limited in range. What enables some attention architectures to retrieve evidence reliably far beyond the training window remains unclear. Through controlled comparisons of positional encodings and attention architectures, we jointly analyze positional generalization, attention entropy stability, and evidence retrieval accuracy to explain the observed extrapolation gaps. First, our probes suggest that a model without explicit positional encoding can still overfit to absolute-position patterns learned within the training window. Second, we find that architectures using the same positional encoding can maintain similarly stable attention entropy yet differ sharply in how highly they rank the required evidence. Third, our comparisons within a single architecture show that changing the positional encoding affects both entropy drift and evidence ranking at longer lengths. These findings explain how positional generalization, attention stability, and evidence selection jointly support reliable retrieval far beyond the training window.
\end{abstract}

\section{Introduction}
\begin{figure*}[t]
\centering
\resizebox{\textwidth}{!}{\input{figures/overview.tex}}
\caption{Overview. The top panel summarizes the analyses of positional generalization, attention entropy stability, and evidence retrieval accuracy. The arrows mark the relationships examined in Secs.~\ref{sec:chunk-position} and~\ref{sec:routing-position}. The bottom panel summarizes the results for each criterion and the length extrapolation performance of 345M models trained with an 8K context. The symbols $\checkmark$, $\sim$, and $\times$ indicate that a criterion is met, partially met, or unmet, respectively. In the Full-HoPE row, the dash indicates that full attention has no block selector.}
\label{fig:overview}
\end{figure*}

Long-document understanding and long-horizon interaction require language models to retrieve information from extended contexts. Length extrapolation enables models trained on shorter sequences to retrieve and use information from longer contexts without further training. One approach to processing very long contexts is to select relevant parts of the key--value cache for each query. Recent methods using this approach have demonstrated strong retrieval performance far beyond the training window~\cite{hu2026hils}. However, retrieval accuracy can still deteriorate sharply beyond the training window, with substantial differences between attention architectures.

Prior work has approached failures beyond the training window in three ways (see Appendix~\ref{app:related-work} for further discussion). The first concerns positional encodings. Beyond the training window, some RoPE components enter phase ranges not encountered during training. HoPE replaces lower-frequency rotary components with position-independent ones~\cite{chen2025hope}, while YaRN combines rotary-frequency interpolation with attention scaling~\cite{peng2024yarn}. Removing the explicit encoding raises a different question about what positional information the model learns on its own. In models without explicit positional encodings (NoPE), probes can recover absolute position within the training window but lose accuracy beyond it~\cite{puvvada2025swan}. The second concerns attention entropy. Prior work links length-generalization failures to increased attention entropy, reflecting a loss of concentration as contexts grow~\cite{wang2024length}. Temperature scaling sharpens attention to counter this effect~\cite{chiang2022limitation,wang2024length}. Sparse attention limits it structurally, since each query attends to only a restricted set of tokens~\cite{vasylenko2026long}. In our experiments, two sparse models using the same positional encoding show similarly small entropy changes, yet one maintains high retrieval accuracy at much greater lengths than the other (Tables~\ref{tab:ruler-long} and~\ref{tab:entropy-drift}).
The third factor, evidence selection, may help explain this gap.  \citet{leng2026length} study how chunk representations and information integration affect length generalization, and find that lower evidence rankings go with lower accuracy. Taken together, these findings link long-context retrieval to positional representations, attention concentration, and evidence ranking. We analyze these factors jointly to characterize the conditions under which models can reliably retrieve evidence from contexts far longer than those seen during training.

First, we vary the positional encoding within HiLS~\cite{hu2026hils} and compare attention architectures under a common positional encoding (Fig.~\ref{fig:overview}).
Both the RoPE and NoPE variants of HiLS lose retrieval accuracy beyond the training window. For RoPE, adjusting rotary frequencies and scaling attention with dynamic YaRN~\cite{peng2024yarn} partially restores retrieval at lengths where some rotary components encounter unfamiliar phases.
In HiLS-NoPE, absolute position becomes easier to decode within 8K as long-context retrieval deteriorates during training, a pattern consistent with positional overfitting. In the second comparison, retrieval accuracy still differs across architectures when the positional encoding is held fixed. As context length increases, more keys compete for attention, which may dilute attention to different degrees across architectures. We therefore investigate whether sparse attention can mitigate this effect by restricting the number of attended tokens, and find that architectures with fixed attention budgets exhibit more stable entropy. To understand why retrieval still differs between the two sparse models with similarly stable entropy, we measure how often each selector ranks a block containing the required evidence first. Across the three probed HiLS configurations, within-window absolute-position decodability is associated with both entropy drift and evidence ranking, linking positional generalization to the other two factors (Fig.~\ref{fig:overview}).

Our contributions are as follows:

\begin{itemize}
    \item \textbf{A joint mechanistic analysis of length extrapolation.}
    We jointly examine three key factors governing reliable length extrapolation: positional generalization, attention entropy stability, and evidence retrieval accuracy, providing a unified perspective on the extrapolation behavior of different attention architectures.

    \item \textbf{Distinct extrapolation failure mechanisms.}
Of the architectures tested with HoPE, only HiLS maintains high task-averaged RULER accuracy through 512K (Table~\ref{tab:ruler-long}). Our diagnostics link the earlier failures of its RoPE and NoPE variants to unfamiliar rotary phases and positional overfitting, respectively (Sec.~\ref{sec:position}). With HoPE fixed, we revisit Native Sparse Attention (NSA)~\cite{yuan2025nsa} and DashAttention~\cite{huang2026dash}, two sparse baselines from the original HiLS RULER study~\cite{hu2026hils}. Full attention and DashAttention exhibit larger entropy drift than HiLS and NSA, which both use fixed block-selection budgets. As context length increases beyond the 8K training window, the rate at which an evidence block ranks first drops sharply in NSA and changes little in HiLS (Tables~\ref{tab:entropy-drift} and~\ref{tab:evidence-ranking}).

    \item \textbf{Mechanistic interactions and inference-time interventions.}
    Across the three probed HiLS configurations, higher within-window absolute-position decodability is associated with both larger entropy drift and poorer evidence ranking at longer lengths. We use dynamic YaRN~\cite{peng2024yarn} to adjust rotary frequencies and attention temperature in RoPE models, and logarithmic query scaling inspired by \citet{chiang2022limitation} to counter attention dilution in full attention.
Both interventions partially restore long-context retrieval.
\end{itemize}

\begin{table*}[!t]
\caption{RULER answer-token exact match (EM, \%) under gold prefixes for
345M models trained at 8K, with 100 examples per task and length.
SN, MK, and VT denote Single-NIAH, Multi-key, and Variable Tracking.
Darker blue indicates higher EM on a 0--100 scale.
Bold and underlined values rank first and second within each column.}
\label{tab:ruler-long}
\centering
\begingroup
\small

\definecolor{rulerBlue}{RGB}{166,199,226}
\setlength{\fboxsep}{0pt}
\setlength{\tabcolsep}{0.5pt}
\renewcommand{\arraystretch}{1.17}

\newcommand{\rulerEM}[2][]{%
  \colorbox{rulerBlue!#2!white}{%
    \makebox[1.77em][c]{%
      \rule[-0.27em]{0pt}{1.12em}%
      \ifnum#2=0\relax
        \textcolor{black!64}{#1{#2}}%
      \else
        #1{#2}%
      \fi
    }%
  }%
}

\begin{tabular}{@{}l@{\hspace{6pt}}ccc*{5}{@{\hspace{6pt}}ccc}@{}}
\toprule[0.65pt]
 & \multicolumn{3}{c}{\textbf{8K}}
 & \multicolumn{3}{c}{\textbf{16K}}
 & \multicolumn{3}{c}{\textbf{32K}}
 & \multicolumn{3}{c}{\textbf{128K}}
 & \multicolumn{3}{c}{\textbf{256K}}
 & \multicolumn{3}{c}{\textbf{512K}} \\
\cmidrule(l{1pt}r{1pt}){2-4}
\cmidrule(l{1pt}r{1pt}){5-7}
\cmidrule(l{1pt}r{1pt}){8-10}
\cmidrule(l{1pt}r{1pt}){11-13}
\cmidrule(l{1pt}r{1pt}){14-16}
\cmidrule(l{1pt}r{1pt}){17-19}
\textbf{Model}
 & SN & MK & VT & SN & MK & VT & SN & MK & VT
 & SN & MK & VT & SN & MK & VT & SN & MK & VT \\
\midrule

Full-HoPE
 & \rulerEM[\textbf]{100}   & \rulerEM[\textbf]{99} & \rulerEM{54}
 & \rulerEM[\underline]{99} & \rulerEM[\textbf]{92} & \rulerEM{16}
 & \rulerEM{75} & \rulerEM{4} & \rulerEM{5}
 & \rulerEM{0}  & \rulerEM{0} & \rulerEM{0}
 & \rulerEM{0}  & \rulerEM{0} & \rulerEM{0}
 & \rulerEM{0}  & \rulerEM{0} & \rulerEM{0} \\

DashAttention-HoPE
 & \rulerEM{85} & \rulerEM{45} & \rulerEM{40}
 & \rulerEM{71} & \rulerEM{18} & \rulerEM{13}
 & \rulerEM{13} & \rulerEM{1}  & \rulerEM{7}
 & \rulerEM{0}  & \rulerEM{0}  & \rulerEM{0}
 & \rulerEM{0}  & \rulerEM{0}  & \rulerEM{0}
 & \rulerEM{0}  & \rulerEM{0}  & \rulerEM{0} \\

NSA-HoPE
 & \rulerEM{89} & \rulerEM{58} & \rulerEM{33}
 & \rulerEM{61} & \rulerEM{24} & \rulerEM{3}
 & \rulerEM{25} & \rulerEM{10} & \rulerEM{5}
 & \rulerEM{9}  & \rulerEM{2}  & \rulerEM{0}
 & \rulerEM{2}  & \rulerEM{1}  & \rulerEM{0}
 & \rulerEM{0}  & \rulerEM{0}  & \rulerEM{0} \\

HiLS-HoPE
 & \rulerEM[\textbf]{100} & \rulerEM{91}          & \rulerEM{73}
 & \rulerEM[\textbf]{100} & \rulerEM{85}          & \rulerEM[\underline]{65}
 & \rulerEM[\textbf]{100} & \rulerEM[\textbf]{89} & \rulerEM[\textbf]{67}
 & \rulerEM[\textbf]{100} & \rulerEM[\textbf]{88} & \rulerEM[\textbf]{69}
 & \rulerEM[\textbf]{99}  & \rulerEM[\textbf]{87} & \rulerEM[\textbf]{63}
 & \rulerEM[\textbf]{99}  & \rulerEM[\textbf]{91} & \rulerEM[\textbf]{52} \\

HiLS-NoPE
 & \rulerEM[\underline]{99} & \rulerEM{91} & \rulerEM[\underline]{77}
 & \rulerEM{92} & \rulerEM{66} & \rulerEM{43}
 & \rulerEM{44} & \rulerEM{16} & \rulerEM{6}
 & \rulerEM{3}  & \rulerEM{0}  & \rulerEM{0}
 & \rulerEM{1}  & \rulerEM{0}  & \rulerEM{0}
 & \rulerEM{0}  & \rulerEM{0}  & \rulerEM{0} \\

HiLS-RoPE
 & \rulerEM[\textbf]{100} & \rulerEM{65} & \rulerEM{53}
 & \rulerEM{0} & \rulerEM{0} & \rulerEM{0}
 & \rulerEM{0} & \rulerEM{0} & \rulerEM{0}
 & \rulerEM{0} & \rulerEM{0} & \rulerEM{0}
 & \rulerEM{0} & \rulerEM{0} & \rulerEM{0}
 & \rulerEM{0} & \rulerEM{0} & \rulerEM{0} \\

HiLS-T5
 & \rulerEM[\textbf]{100}   & \rulerEM[\underline]{95} & \rulerEM[\textbf]{81}
 & \rulerEM[\textbf]{100}   & \rulerEM[\underline]{90} & \rulerEM[\textbf]{73}
 & \rulerEM[\underline]{99} & \rulerEM[\underline]{88} & \rulerEM[\underline]{59}
 & \rulerEM[\underline]{97} & \rulerEM[\underline]{64} & \rulerEM[\underline]{35}
 & \rulerEM[\underline]{92} & \rulerEM[\underline]{57} & \rulerEM[\underline]{30}
 & \rulerEM[\underline]{96} & \rulerEM[\underline]{66} & \rulerEM[\underline]{27} \\
\bottomrule[0.65pt]
\end{tabular}
\endgroup
\end{table*}
\section{Background}
\label{sec:background}

\subsection{Positional Encodings and Length Generalization}

Since self-attention does not inherently encode token order, positional information is important for modeling sequential relationships. RoPE~\cite{su2024roformer} encodes relative positions through rotations of queries and keys at multiple frequencies. In Gemma 7B, for example, some attention heads use high-frequency RoPE components to attend to the current or previous token~\cite{barbero2025round}. While HoPE~\cite{chen2025hope} retains high-frequency rotary components and replaces lower-frequency components with position-independent ones. T5~\cite{raffel2020t5} introduces learned, bucketed relative-position biases into attention logits.

Positional encodings can impair length extrapolation in two ways. The first is out-of-distribution positions. Beyond the training window, some rotary components enter phase ranges not encountered during training, potentially changing attention patterns~\cite{liu2024scaling,chen2025hope}. To reduce this mismatch, Position Interpolation down-scales position indices to stay within the trained range~\cite{chen2023pi}. YaRN~\cite{peng2024yarn} combines RoPE frequency interpolation with attention temperature scaling to extend the context window.

The second is positional overfitting. NoPE omits explicit positional encodings but can learn positional information implicitly during training~\cite{haviv2022nope}. Prior probe analyses find that absolute-position mappings decoded from NoPE hidden states can fail beyond the training window~\cite{puvvada2025swan}. Such failures motivate studying positional overfitting, in which learned positional dependencies are specific to the training window.

\subsection{Attention Entropy and Sparse Attention}

As context grows, additional keys can spread attention away from relevant information, increasing attention entropy. Under bounded logits and fixed temperature, softmax loses sharpness as the number of competing keys grows~\cite{velickovic2025softmax}: attention entropy grows without limit as the sequence becomes longer~\cite{han2024lminfinite}, and the largest attention weight approaches zero, a phenomenon termed attention fading~\cite{nakanishi2025scalable}. Empirical work also links NoPE's extrapolation failures to attention dispersion~\cite{wang2024length}.

One way to counter this dispersion is to sharpen the attention distribution at inference time. Logarithmic query scaling, related to log-length attention scaling~\cite{chiang2022limitation}, adjusts query magnitudes as context length increases to sharpen attention distributions without modifying positional frequencies.

Another way to limit dispersion is to restrict which tokens receive nonzero attention~\cite{vasylenko2026long}. In chunk-based sparse attention, the context is partitioned into blocks, and a selection mechanism identifies relevant blocks for subsequent token-level attention. The architectures differ primarily in how they represent, score, and select these blocks.


HiLS~\cite{hu2026hils} uses landmark tokens to retrieve up to $K$ chunks per query and combines their within-chunk attention outputs using retrieval scores. Native Sparse Attention (NSA)~\cite{yuan2025nsa} also uses a fixed block budget. It scores blocks using compressed representations and combines compressed, selected-token, and local attention through learned gates. DashAttention~\cite{huang2026dash} uses $\alpha$-entmax to select a variable number of blocks, with the nonzero weights serving as a prior for token-level attention.

\section{Length Extrapolation across Attention Architectures}
\label{sec:setup}
\label{sec:task-results}

We study 345M-parameter decoder-only Transformers trained with an 8K context. To disentangle the effects of positional encoding and attention architecture, we conduct two controlled comparisons: varying positional encodings (HoPE, NoPE, RoPE, and T5) under a fixed HiLS architecture, and varying attention architectures (HiLS, full attention, NSA, and DashAttention) under a fixed HoPE encoding. We evaluate three RULER tasks~\cite{hsieh2024ruler}---Single-NIAH (SN), Multi-key (MK), and Variable Tracking (VT)---at context lengths from 8K to 512K. Training and evaluation details are provided in Appendix~\ref{app:training-settings}.

The two comparisons reveal striking differences in length extrapolation (Table~\ref{tab:ruler-long}). Under the same HiLS architecture, RoPE fails at 16K, the first length we evaluate beyond the 8K training window, NoPE degrades sharply by 32K, and T5 maintains strong Single-NIAH accuracy through 512K but deteriorates on Multi-key and Variable Tracking. In contrast, HiLS-HoPE retains at least 80\% of its 8K task-averaged EM through 512K. Even with HoPE fixed, changing the attention architecture produces similarly large differences: full attention, NSA, and DashAttention lose retrieval accuracy at substantially shorter context lengths than HiLS.

These results raise two questions: why do positional encodings lead to such different extrapolation behavior under the same attention architecture, and why do attention architectures differ so substantially even with the same positional encoding? We address the first question by examining how positional representations generalize beyond the training window (Sec.~\ref{sec:position}). For the second, we measure how attention entropy changes as the context grows (Sec.~\ref{sec:chunk}) and whether sparse selectors continue to rank blocks containing the required evidence highly (Sec.~\ref{sec:routing}).

\section{Positional Mechanisms beyond the Training Window}\label{sec:position}

Within the same HiLS architecture, different positional encodings lead to markedly different extrapolation behavior: RoPE fails at 16K, NoPE degrades sharply by 32K, while HoPE maintains strong retrieval at much longer lengths (Table~\ref{tab:ruler-long}). These differences motivate us to investigate why both explicit positional encoding (RoPE) and the absence of explicit positional encoding (NoPE) can lead to extrapolation failures.

\begin{figure*}[t]
\centering
\includegraphics[width=\textwidth]{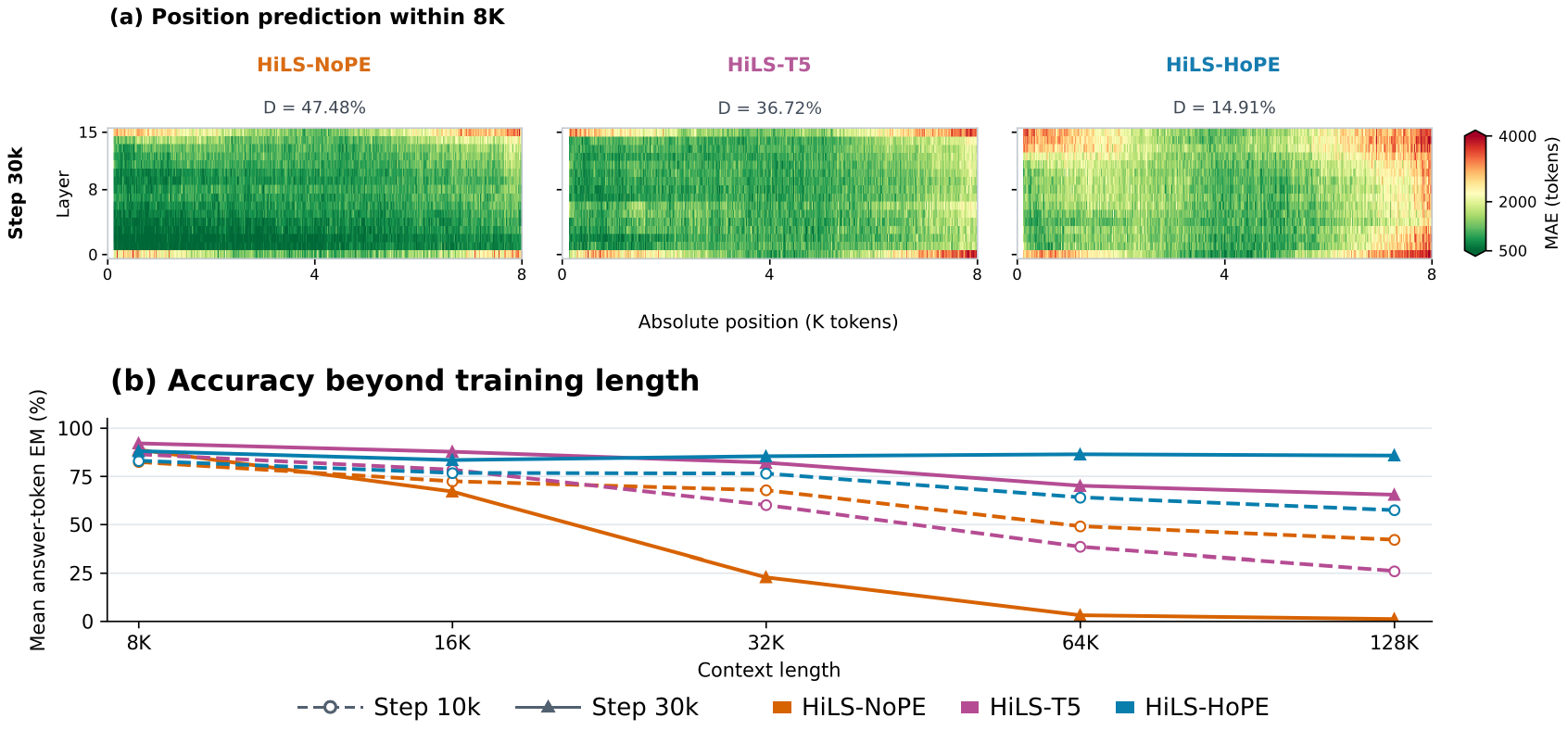}
\caption{(a) Within-window position-probe MAE of HiLS at step 30k (lower is more decodable). (b) Mean RULER EM at steps 10k (dashed) and 30k (solid).}
\label{fig:nope-training-checkpoints}
\vspace{4pt}
\centering
\includegraphics[width=.85\textwidth]{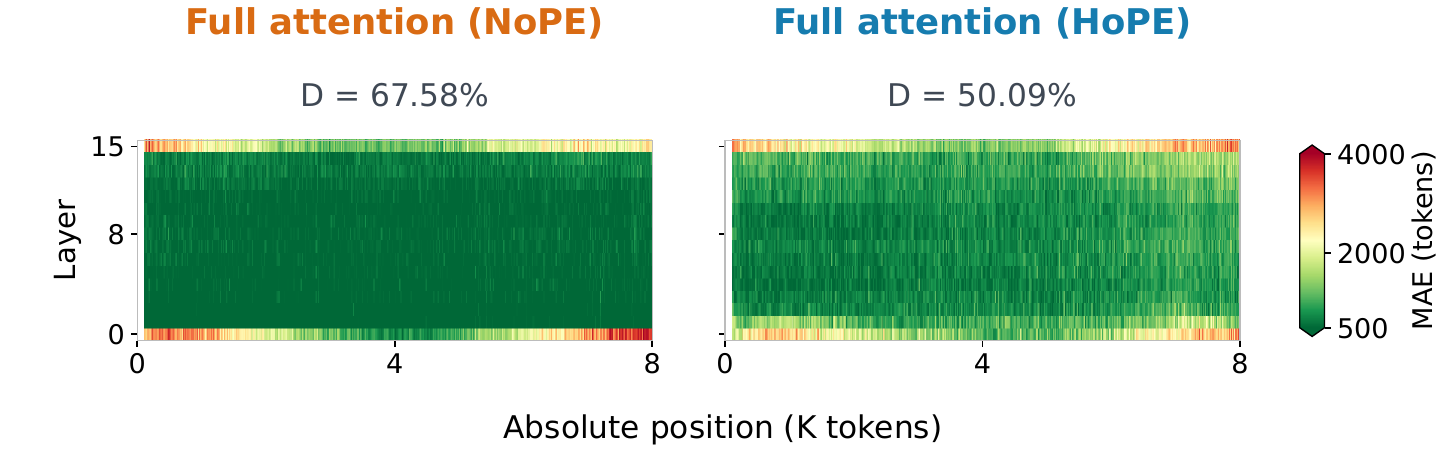}
\caption{Within-8K position-probe MAE for Full-NoPE at step 30k (left) and Full-HoPE at step 30k (right). Lower MAE (green) means more decodable absolute position.}
\label{fig:dense-probe}
\end{figure*}

    \subsection{RoPE Extrapolation and Unseen Rotary Phases}\label{sec:pos-explicit}

As context length exceeds the training window, the low-frequency rotary components of RoPE can encounter relative phases not observed during training~\cite{su2024roformer,chen2025hope}. HoPE replaces these low-frequency RoPE dimensions with NoPE dimensions, retaining rotary encoding only in the high-frequency dimensions~\cite{chen2025hope}. This difference offers a possible explanation for why HiLS-RoPE fails beyond 8K while HiLS-HoPE maintains strong retrieval (Table~\ref{tab:ruler-long}).

To investigate whether inference-time adjustments can mitigate this failure, we apply dynamic YaRN~\cite{peng2024yarn} to HiLS-RoPE and Full-RoPE without retraining. We set the dynamic scaling factor to $s=L/8192$ for each evaluated context length $L$ (see the full YaRN configuration in Appendix~\ref{app:yarn}). At 8K, accuracy remains essentially unchanged. Beyond 8K, YaRN substantially restores Single-NIAH and Multi-key performance in both models, although recovery weakens at longer lengths and remains limited on Variable Tracking (Table~\ref{tab:interventions}). These results suggest that unseen rotary phases contribute to RoPE's extrapolation failure and that adjusting positional encoding at inference time can substantially improve length extrapolation.

\begin{table}[t]
\caption{RULER answer-token EM (\%, SN/MK/VT) for RoPE models with and without dynamic YaRN~\cite{peng2024yarn}. For each model, both runs use the same checkpoint weights and the RULER prompts from Table~\ref{tab:ruler-long}. Appendix~\ref{app:yarn} gives the YaRN configuration.}
\label{tab:interventions}
\centering
\footnotesize
\setlength{\tabcolsep}{2pt}
\resizebox{\columnwidth}{!}{\begin{tabular}{@{}lccccc@{}}
\toprule
Model & 8K & 16K & 32K & 64K & 128K \\
\midrule
Full-RoPE & 100/99/42 & 0/0/0 & 0/0/0 & 0/0/0 & 0/0/0 \\
\quad + YaRN & 100/99/42 & 99/93/21 & 92/84/6 & 74/58/3 & 44/39/1 \\
HiLS-RoPE & 100/65/54 & 0/0/0 & 0/0/0 & 0/0/0 & 0/0/0 \\
\quad + YaRN & 100/65/53 & 100/65/20 & 96/60/10 & 81/42/12 & 71/27/2 \\
\bottomrule
\end{tabular}}
\end{table}

\subsection{NoPE Extrapolation and Implicit Positional Overfitting}
\label{sec:pos-implicit}
\label{sec:pos-transfer}

HiLS-NoPE has no explicit positional encoding, yet it also fails to extrapolate. Prior work showed that probes can recover absolute position from NoPE hidden states within the training window but can lose accuracy beyond it~\cite{puvvada2025swan}. We build on this observation by asking whether absolute position becomes more decodable as HiLS-NoPE's long-context retrieval deteriorates during training. We also examine whether probe accuracy and retrieval performance decline over a similar range of context lengths at the final checkpoint. We fit a separate two-layer MLP probe to each decoder layer's hidden states to predict absolute token positions within 8K. We define decodability $D$ as the percentage reduction in mean absolute error relative to a constant-position predictor, averaged across layers (Appendix~\ref{app:position-decodability}).

\begin{figure*}[t]
\centering
    \includegraphics[width=\textwidth]
    {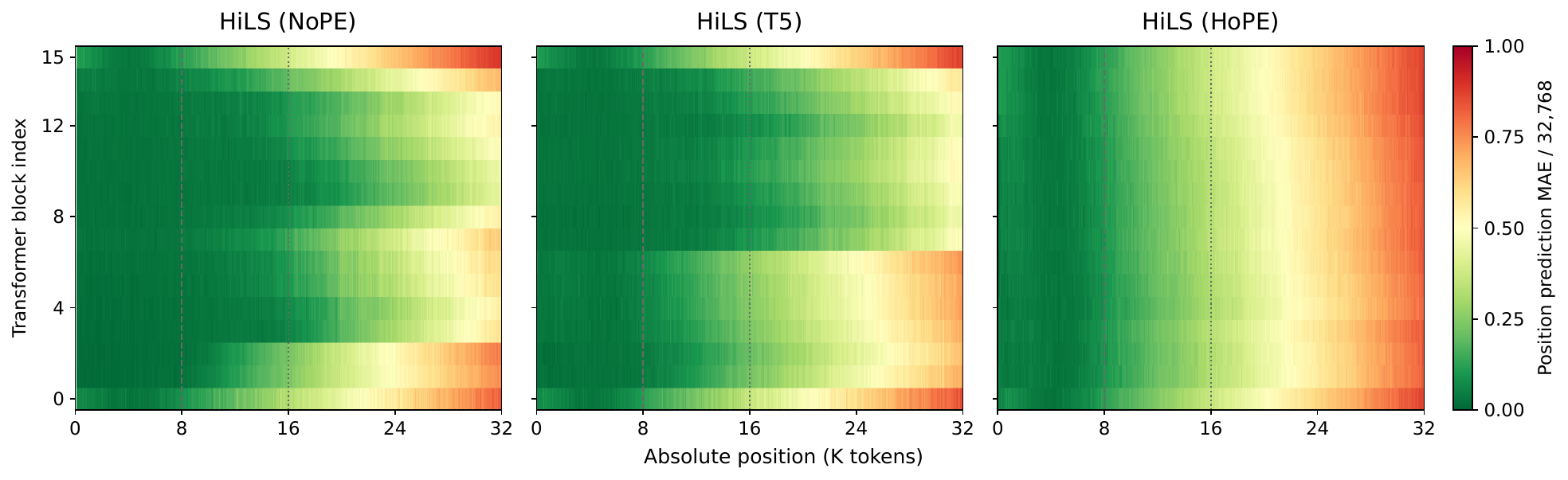}
\caption{Absolute-position error of frozen two-layer MLP probes for HiLS-NoPE, HiLS-T5, and HiLS-HoPE at 30{,}000 pretraining step, fitted within 8K and evaluated on held-out 32K prefixes. Colors show MAE divided by 32{,}768 on a shared scale (green is lower). Dashed and dotted lines mark 8K and 16K, respectively.}
    \label{fig:frozen-position-heatmap-345m}
\end{figure*}

After 30k pretraining steps, the within-window position decodability score $D$ is about three times as high for HiLS-NoPE as for HiLS-HoPE (Fig.~\ref{fig:nope-training-checkpoints}a). We observe the same ordering under full attention, where absolute position is more decodable in Full-NoPE than in Full-HoPE (Fig.~\ref{fig:dense-probe}).

Between 10k and 30k pretraining steps, $D$ increases by 7.19 percentage points for HiLS-NoPE and by only 0.63 points for HiLS-HoPE (Table~\ref{tab:checkpoint-position}). As absolute position becomes more decodable in HiLS-NoPE, its mean RULER answer-token EM rises from 82.3\% to 88.7\% at 8K but falls from 42.0\% to 1.0\% at 128K (Fig.~\ref{fig:nope-training-checkpoints}b). This combination of stronger within-window position information and poorer long-context retrieval is consistent with positional overfitting.

\begin{table}[t]
\caption{Within-window position decodability $D$ over decoder blocks 0--15 at 10k and 30k pretraining steps. Changes are in percentage points (pp), computed before rounding.}
\label{tab:checkpoint-position}
\centering
\small
\begin{tabular}{@{}lrrr@{}}
\toprule
Model & 10k $D$ & 30k $D$ & Change \\
\midrule
HiLS-NoPE & 40.29\% & 47.48\% & $+7.19$ pp \\
HiLS-T5 & 33.42\% & 36.72\% & $+3.30$ pp \\
HiLS-HoPE & 14.28\% & 14.91\% & $+0.63$ pp \\
\bottomrule
\end{tabular}
\end{table}



We next test how far the position readouts fitted within 8K transfer beyond the training window. For HiLS-NoPE, HiLS-T5, and HiLS-HoPE, we fit a probe to each layer of the final checkpoint using positions within 8K and evaluate the frozen probes on held-out 32K prefixes (Fig.~\ref{fig:frozen-position-heatmap-345m}). For HiLS-NoPE and HiLS-T5, intermediate-layer probes retain low position error up to roughly 16K, with errors increasing toward 32K. HiLS-NoPE's retrieval accuracy follows a similar pattern over this range, remaining substantial at 16K before declining sharply at 32K (Table~\ref{tab:ruler-long}).

In HiLS-HoPE, by comparison, within-window decodability stays low at both checkpoints (Table~\ref{tab:checkpoint-position}), and retrieval remains strong at 32K and beyond (Table~\ref{tab:ruler-long}). For HiLS-NoPE, the limited transfer of the frozen position readouts lends further support to positional overfitting as an explanation for its decline in long-context retrieval accuracy.

\section{Attention Dilution across Context Lengths}\label{sec:chunk}

HiLS-HoPE maintains strong retrieval far beyond the 8K training window (Sec.~\ref{sec:position}). We therefore use HoPE as the common positional encoding to compare attention architectures. Even with HoPE fixed, full attention, NSA, DashAttention, and HiLS still exhibit substantially different length extrapolation performance (Table~\ref{tab:ruler-long}). Prior work links extrapolation failures to attention dilution as more keys compete for attention~\cite{wang2024length} and shows that restricting the set of attended tokens can limit this dispersion~\cite{vasylenko2026long}. We ask whether the architectures that lose retrieval accuracy earlier also exhibit larger changes in attention entropy.

\subsection{Entropy Stability and Retrieval Performance}
\label{sec:ent-measure}
\label{sec:ent-structure}

Attention dilution can manifest as changes in the concentration of attention distributions as context length increases. To quantify this behavior across architectures, we measure \emph{full-context entropy} using an effective attention weight $a_{L,t}$ per raw token. For a query at zero-based position $p$ in an input of length $L$, the full-context entropy is
\begin{equation}
H_L(a)=-\sum_{t=0}^{p}a_{L,t}\log a_{L,t}.
\end{equation} We report the mean absolute entropy change relative to 8K, comparing the same query tokens across context lengths. Details of the effective attention weights and paired evaluation are provided in Appendix~\ref{app:entropy-protocol}.

With HoPE fixed, full attention and DashAttention exhibit substantially larger entropy drift than NSA and HiLS (Table~\ref{tab:entropy-drift}). This drift coincides with rapid retrieval degradation, whereas HiLS maintains relatively stable entropy and strong retrieval at longer lengths. NSA, however, also exhibits stable entropy despite poor extrapolation performance, suggesting that attention stability alone cannot explain all retrieval failures.

\begin{table}[t]
\caption{Full-context entropy drift relative to 8K. Entries are mean
absolute entropy changes in nats, measured on matched query tokens
across 100 passages and averaged over layers and heads. The Extrap.
column summarizes RULER extrapolation as in Fig.~\ref{fig:overview}
($\checkmark$ strong, $\sim$ partial, $\times$ limited).
The dash marks the inference-time scaling variant.}
\label{tab:entropy-drift}
\centering
\small
\setlength{\tabcolsep}{2pt}
\renewcommand{\arraystretch}{1.08}
\begin{tabular*}{\columnwidth}{@{\extracolsep{\fill}}lcrrr@{}}
\toprule
 & & \multicolumn{3}{c}{Entropy drift from 8K} \\
\cmidrule(lr){3-5}
Model & Extrap. & 16K & 32K & 64K \\
\midrule
Full-HoPE          & \extrapLimited & 0.709 & 1.664 & 2.635 \\
\quad + log scaling & ---           & 0.360 & 0.850 & 1.310 \\
\addlinespace[3pt]
DashAttention-HoPE & \extrapLimited & 0.510 & 1.026 & 1.381 \\
NSA-HoPE           & \extrapLimited & 0.126 & 0.223 & 0.313 \\
HiLS-HoPE          & \extrapStrong  & 0.099 & 0.154 & 0.201 \\
\addlinespace[3pt]
HiLS-RoPE          & \extrapLimited & 2.367 & 1.669 & 1.725 \\
HiLS-NoPE          & \extrapLimited & 0.675 & 1.361 & 1.453 \\
HiLS-T5            & \extrapPartial & 0.377 & 0.644 & 0.807 \\
\bottomrule
\end{tabular*}
\end{table}
In Full-HoPE, mean full-context entropy rises from 4.44 nats at 8K to 7.07 nats at 64K. We therefore test whether sharpening attention at inference time improves retrieval beyond the training window. Scaling queries by a factor greater than one amplifies differences between attention logits and sharpens the softmax distribution. Inspired by log-length attention scaling~\cite{chiang2022limitation}, we scale the normalized queries before rotary encoding, $Q'_p=\alpha(p)\,Q_p$, with
\begin{equation}
\alpha(p)=1+0.1\ln\!\left(\max\!\left(\tfrac{p+1}{8192},1\right)\right),
\label{eq:log-query-scaling}
\end{equation}
where $p$ is the zero-based query position. Queries within the first 8,192 positions are unchanged. At inference time, logarithmic query scaling reduces entropy drift in Full-HoPE by about 50\% across the measured lengths and largely restores Single-NIAH and Multi-key performance at 32K (Table~\ref{tab:entropy-drift}; Appendix~\ref{app:concentration-adjustment}). These results indicate that attention dilution is an important bottleneck for full attention during length extrapolation, and that improving attention concentration can substantially mitigate its retrieval degradation.

\subsection{Influence of Positional Encoding on Entropy Stability}
\label{sec:chunk-position}

\begin{figure}[t]
\centering
\includegraphics[width=\columnwidth,trim=0 0 357 20,clip]{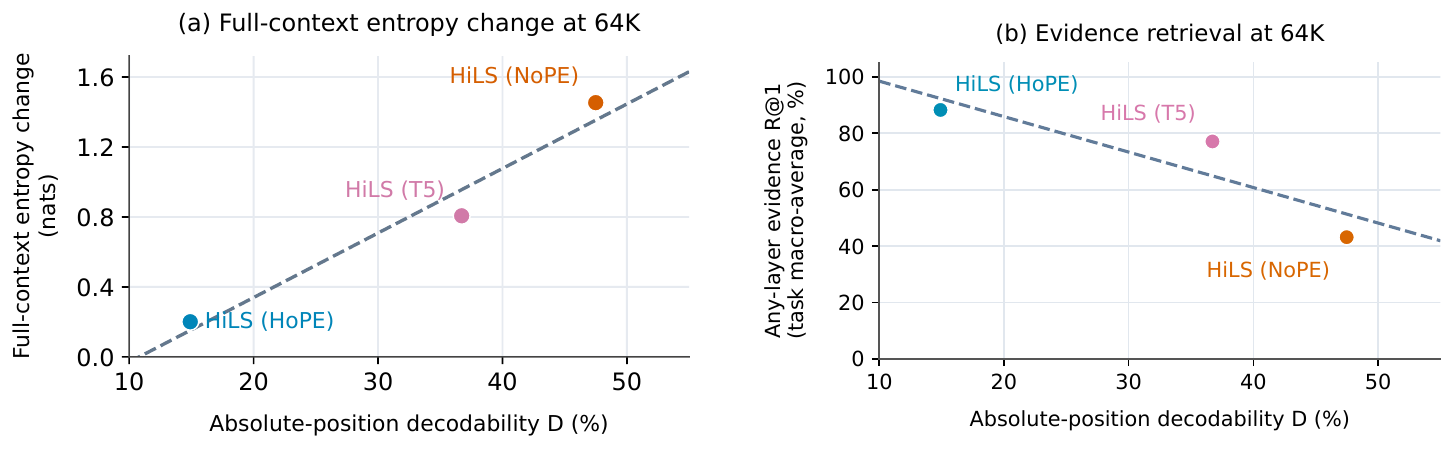}
\caption{Higher within-window position decodability is associated with larger entropy change from 8K to 64K across the three probed HiLS settings (dashed: linear fit).}
\label{fig:decodability-entropy}
\end{figure}

With HoPE fixed, entropy drift differs substantially across attention architectures. We now compare positional encodings within HiLS, keeping the architecture and its block-selection budget fixed. At 64K, HiLS-NoPE and HiLS-RoPE exhibit seven to nine times larger entropy drift than HiLS-HoPE (Table~\ref{tab:entropy-drift}). Notably, HiLS-RoPE already shows substantial entropy drift at 16K, where its retrieval accuracy collapses. These results show that a fixed sparse attention structure does not guarantee entropy stability across positional encodings. Furthermore, across the three probed HiLS configurations (NoPE, T5, and HoPE), higher within-window position decodability is associated with larger entropy drift (Fig.~\ref{fig:decodability-entropy}). This association suggests that positional representations may influence attention concentration through changes in query--key scores, linking positional generalization to attention entropy stability.

\begin{table*}[t]
\caption{Evidence ranking on RULER. Entries are the percentages of gold
evidence spans overlapping a block ranked first (R@1) or among the top
eight (Coverage@8) in at least one layer and routing group. All models
use the same 100 prompts per task and length. The Extrap. column summarizes RULER extrapolation for the task shown in each panel ($\checkmark$ strong, $\sim$ partial,
$\times$ limited).}
\label{tab:evidence-ranking}
\centering
\small
\setlength{\tabcolsep}{3pt}
\renewcommand{\arraystretch}{1.04}
\begin{tabular*}{\textwidth}
  {@{\extracolsep{\fill}}lcrrrr@{\hspace{1.6em}}rrrr@{}}
\toprule
 & 
 & \multicolumn{4}{c}{\textbf{Any-layer R@1 (\%)}} 
 & \multicolumn{4}{c}{Coverage@8 (\%)} \\
\cmidrule(lr){3-6}\cmidrule(lr){7-10}
Model & Extrap. & 8K & 16K & 32K & 64K & 8K & 16K & 32K & 64K \\
\midrule

\multicolumn{10}{@{}l}{\textit{Single-NIAH}} \\
DashAttention-HoPE & \extrapLimited & 38.0 & 22.0 &   6.0 &  1.0 & 90.0 & 58.0 &  22.0 & 14.0 \\
NSA-HoPE           & \extrapLimited & 29.0 & 26.0 &   6.0 & 10.0 & 84.0 & 62.0 &  35.0 & 34.0 \\
HiLS-HoPE          & \extrapStrong  & 97.0 & 95.0 & 100.0 & 98.0 & 97.0 & 95.0 & 100.0 & 98.0 \\
HiLS-RoPE          & \extrapLimited & 97.0 &  8.0 &   5.0 &  2.0 & 97.0 & 32.0 &  16.0 & 18.0 \\
HiLS-NoPE          & \extrapLimited & 97.0 & 94.0 &  91.0 & 74.0 & 97.0 & 95.0 & 100.0 & 96.0 \\
HiLS-T5            & \extrapStrong & 97.0 & 95.0 & 100.0 & 98.0 & 97.0 & 95.0 & 100.0 & 98.0 \\

\midrule[0.3pt]
\multicolumn{10}{@{}l}{\textit{Multi-key}} \\
DashAttention-HoPE & \extrapLimited & 29.0 & 23.5 &  7.0 &  3.0 & 82.5 & 57.5 & 29.5 &  14.5 \\
NSA-HoPE           & \extrapLimited & 32.0 & 17.0 &  7.0 &  5.5 & 78.5 & 60.0 & 40.5 &  26.5 \\
HiLS-HoPE          & \extrapStrong  & 91.0 & 94.5 & 96.5 & 95.5 & 93.5 & 97.5 & 99.5 & 100.0 \\
HiLS-RoPE          & \extrapLimited & 71.5 &  5.0 &  3.0 &  2.5 & 93.5 & 24.5 & 11.5 &  13.5 \\
HiLS-NoPE          & \extrapLimited & 78.5 & 65.5 & 39.0 & 27.5 & 93.5 & 97.5 & 97.5 &  96.0 \\
HiLS-T5            & \extrapPartial & 74.0 & 79.5 & 84.0 & 76.0 & 93.5 & 97.5 & 99.5 & 100.0 \\

\midrule[0.3pt]
\multicolumn{10}{@{}l}{\textit{Variable Tracking}} \\
DashAttention-HoPE & \extrapLimited & 17.7 & 14.7 & 10.7 &  2.7 & 74.0 & 50.7 & 29.3 & 12.7 \\
NSA-HoPE           & \extrapLimited & 25.7 & 17.0 &  8.7 &  5.0 & 74.0 & 62.0 & 41.7 & 25.3 \\
HiLS-HoPE          & \extrapStrong  & 69.7 & 65.7 & 68.3 & 71.3 & 96.0 & 96.7 & 98.3 & 99.0 \\
HiLS-RoPE          & \extrapLimited & 54.3 &  8.0 &  1.7 &  1.3 & 95.0 & 29.7 & 15.7 & 13.0 \\
HiLS-NoPE          & \extrapLimited & 46.3 & 54.7 & 38.7 & 28.0 & 96.0 & 96.7 & 96.0 & 90.7 \\
HiLS-T5            & \extrapPartial & 35.0 & 39.7 & 48.3 & 57.3 & 96.0 & 96.7 & 98.3 & 99.3 \\
\bottomrule
\end{tabular*}
\end{table*}

\section{Evidence Selection in Sparse Attention}\label{sec:routing}

Despite exhibiting entropy stability comparable to HiLS, NSA suffers substantial retrieval degradation beyond the 8K training window under the same HoPE positional encoding (Table~\ref{tab:ruler-long}; Table~\ref{tab:entropy-drift}). This discrepancy suggests that stable attention entropy alone cannot explain reliable length extrapolation. We therefore investigate whether differences in evidence selection and ranking account for the remaining performance gap.

\subsection{Evidence Retrieval across Sparse Attention Mechanisms}
\label{sec:sel-measure}
\label{sec:sel-structure}
\citet{leng2026length} show that lower evidence rankings are associated with reduced answer accuracy. For each query, the model's selector ranks candidate blocks of text at each decoder layer, producing one list per \emph{routing group} (one or more attention heads that use the same ranking). Any-layer R@1 is the fraction of gold evidence spans that overlap the highest-ranked block on at least one such list across layers. Coverage@8 is the corresponding fraction for the top eight blocks. Formal definitions and evaluation details are provided in Appendix~\ref{app:depth-selection}.

With HoPE fixed, HiLS exhibits substantially higher evidence R@1 than NSA and DashAttention. At 8K, HiLS-HoPE achieves 69.7--97.0\% R@1 across the three tasks. On every task, NSA and DashAttention both achieve at least 74\% Coverage@8, but neither exceeds 38\% R@1 (Table~\ref{tab:evidence-ranking}). This gap indicates that the baselines often rank the required evidence among their leading candidates but often fail to assign it the highest priority.

As context length increases, NSA and DashAttention exhibit further declines in evidence selection accuracy. By 64K, their R@1 falls to at most 10\%, accompanied by declining Coverage@8, whereas HiLS-HoPE largely preserves its evidence ranking accuracy (Table~\ref{tab:evidence-ranking}). These trends broadly align with their RULER extrapolation performance. Notably, NSA loses evidence recall and retrieval accuracy despite maintaining relatively stable attention entropy (Table~\ref{tab:ruler-long}; Table~\ref{tab:entropy-drift}). This discrepancy shows that attention stability alone cannot explain reliable extrapolation, and that accurate evidence selection constitutes an additional bottleneck.

\begin{figure}[!t]
\centering
\includegraphics[
  width=\columnwidth,
  trim=340 0 0 26,
  clip
]{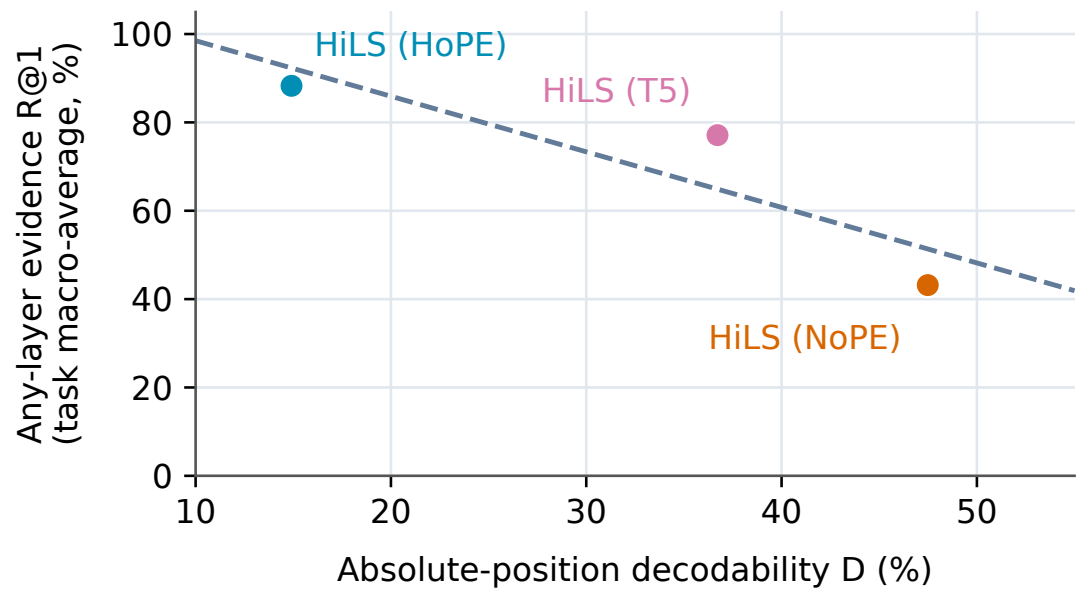}
\caption{Higher within-window position decodability is associated with lower 64K evidence R@1 (averaged over tasks) across the three probed HiLS settings (dashed: linear fit).}
\label{fig:decodability-r1}
\end{figure}

\subsection{Influence of Positional Encoding on Evidence Retrieval}
\label{sec:routing-position}

While evidence ranking depends on the selection mechanism, positional encoding may also influence how candidate chunks are scored as context length grows. To examine this, we compare different positional encodings under the same HiLS selector. At 8K, HiLS-RoPE and HiLS-NoPE already exhibit lower evidence R@1 than HiLS-HoPE on Multi-key and Variable Tracking. By 64K, RoPE loses nearly all top-ranked evidence, while NoPE also suffers substantial ranking degradation on these tasks (Table~\ref{tab:evidence-ranking}). Notably, NoPE still retains 90.7--96.0\% of the required evidence among its top eight candidates despite the decline in R@1, suggesting that its failure involves inaccurate evidence prioritization rather than simply excluding the evidence from the candidate set. Across the three probed HiLS configurations (NoPE, T5, and HoPE), higher within-window absolute-position decodability is associated with lower R@1 at 64K (Fig.~\ref{fig:decodability-r1}). Together, these results indicate that positional encoding influences not only attention entropy stability but also evidence ranking, linking positional generalization to both downstream mechanisms.

\section{Conclusion}
\label{sec:conclusion}

HiLS-HoPE retains most of its 8K task-averaged RULER accuracy through 512K. To understand why other models trained with the same context length lose accuracy sooner, we varied the positional encoding within HiLS and compared attention architectures under HoPE.

\begin{itemize}
    \item \textbf{Unseen rotary phases and positional overfitting can limit length extrapolation.}
    HiLS-RoPE loses accuracy as some rotary components enter unfamiliar phase ranges, and dynamic YaRN partially restores retrieval. In HiLS-NoPE, continued training makes absolute position easier to decode within the training window even as long-context retrieval deteriorates. Probes fitted within that window also lose accuracy at longer lengths.

    \item \textbf{Attention dilution contributes to long-context retrieval failures in full attention.}
    As context grows, more keys compete in full attention, and DashAttention can select more blocks per query. With HoPE fixed, both exhibit larger entropy drift than NSA and HiLS, whose selectors use fixed block budgets. In Full-HoPE, logarithmic query scaling at inference time reduces this drift and partly restores retrieval beyond the training window.

    \item \textbf{Evidence ranking can deteriorate even when attention entropy changes little.}
    Even with fixed selection budgets, NSA and HiLS face a growing pool of candidate blocks. Under HoPE, both maintain relatively stable entropy, yet NSA loses RULER accuracy at shorter lengths. At longer lengths, NSA rarely ranks a block containing the required evidence first, while HiLS largely preserves its within-window ranking accuracy.
\end{itemize}

The three probed HiLS variants share the same selector design and block budget. Across these variants, stronger within-window absolute-position decodability is associated with both larger entropy drift and poorer evidence ranking at longer lengths. These findings suggest that positional encodings, attention stability, and evidence selection should be evaluated together when analyzing and designing models for length extrapolation.

\section*{Limitations}

Among the evaluated configurations, only HiLS-HoPE satisfies all three conditions jointly. Further comparisons are needed to examine whether these conditions also characterize successful extrapolation in other architectures. Our experiments use 345M-parameter models. The associations of decodability with entropy drift and evidence ranking are based on three HiLS configurations. Extending these comparisons to larger models and to natural-document benchmarks such as LongBench~\cite{bai2024longbench} remains future work.

\bibliography{references}
\clearpage
\appendix
\raggedbottom
\makeatletter
\setlength{\@dblfptop}{0pt}
\setlength{\@dblfpbot}{0pt plus 1fil}
\makeatother
\section{Training and Evaluation Details}
\label{app:training-settings}

\subsection{Model Architectures}
\label{sec:models}

We study 345M-parameter decoder-only Transformers. HiLS retrieves 32 landmark-indexed chunks per layer alongside a 512-token local window. HiLS-T5 adds T5's layer-shared relative position bias~\cite{raffel2020t5} to the token-level attention logits of HiLS-NoPE. The relative bias uses 32 buckets with a maximum bucketing distance of 128. HiLS-RoPE uses a base of 10{,}000. HiLS-HoPE uses the same base, retaining rotary components with periods of at most 4{,}096 tokens and setting the remaining angular frequencies to zero. HiLS and NSA use fixed chunk budgets, while DashAttention can attend to more chunks as context grows. We additionally include Full-NoPE in the within-window position-probe comparison.

Each HiLS chunk contains 63 raw tokens and one landmark.
NSA uses a nominal 2K-token retrieval budget.
\subsection{Training Configuration}

Following \citet{hu2026hils}, we train at 8K on approximately 30B tokens from Dolma~\cite{soldaini2024dolma}, converting 5\% of samples into RULER-style retrieval tasks. For the models we train, we use step-30k checkpoints for all analyses unless stated otherwise. Table~\ref{tab:small-scale-training} lists the architecture and training settings of these models.

\begin{table}[!htbp]
\caption{Architecture and training settings of the 345M models.}
\label{tab:small-scale-training}
\centering
\small
\setlength{\tabcolsep}{3pt}
\begin{tabular}{ll}
\toprule
Hyperparameter & 345M setting \\
\midrule
Parameter scale & 345M \\
Layers & 16 \\
Hidden size & 1024 \\
FFN size & 4096 \\
Query / KV heads & 16 / 2 \\
Training context & 8192 tokens \\
Global batch size & 128 \\
Optimizer steps & 30{,}000 \\
Optimizer & AdamW \\
HiLS Local sliding window & 512 tokens \\
HiLS chunk size & 64 \\
HiLS top-$K$ & 32 \\
HiLS LoRA-Q bottleneck & 64 \\
Base vocabulary & 100{,}278 \\
Additional HiLS embedding & 1 landmark token \\
Learning rate & $3\times10^{-4}$ \\
Learning-rate warmup ratio & 0.007 \\
Learning-rate schedule & Constant \\
Weight decay & 0.01 \\
Gradient clipping norm & 1.0 \\
\bottomrule
\end{tabular}
\end{table}

\subsection{Evaluation Protocol}
\label{sec:eval}

We evaluate Single-NIAH (SN), Multi-key (MK), and Variable Tracking (VT) from RULER~\cite{hsieh2024ruler} on 100 examples per task and length from 8K to 512K. We follow the RULER data generation and evaluation implementation in HiLS ~\cite{hu2026hils}. Answer-token exact match (EM) requires every answer token, excluding the terminal EOS token, to be predicted correctly under gold prefixes. We define \emph{reach} as the longest evaluated length at which EM retains at least 80\% of the model's own 8K EM, computed per task or on the task average. Each analysis uses matched inputs across models. All position-probe, entropy, and evidence-retrieval analyses use decoder blocks 0--15.

\section{Positional Mechanisms beyond the Training Window}
\label{app:frozen-position}

\subsection{YaRN Configuration}
\label{app:yarn}
We apply inference-only dynamic YaRN~\cite{peng2024yarn} to step-30k HiLS-RoPE and Full-RoPE. We use $s=L/8192$, $\beta_{\text{fast}}=32$, $\beta_{\text{slow}}=1$, and attention factor $1+0.1\ln s$, preserving the original RoPE base. For each model, baseline and YaRN runs use identical checkpoint weights and the RULER prompts from Table~\ref{tab:ruler-long}.

\subsection{Within-Window Position Probes and Decodability}
\label{app:position-decodability}
For the HiLS checkpoints at steps 10k and 30k, we fit a two-layer MLP probe (Table~\ref{tab:position-probe-settings}) on each layer's hidden states to predict token position within 8K and evaluate its mean absolute error $\mathrm{MAE}_\ell$ on held-out sequences. 
\begin{table}[t]
\caption{Position-probe architecture and fitting settings.}
\label{tab:position-probe-settings}
\centering
\small
\setlength{\tabcolsep}{5pt}
\renewcommand{\arraystretch}{1.1}
\begin{tabular}{@{}ll@{}}
\toprule
Setting & Value \\
\midrule
Probe architecture & Two-layer MLP \\
Hidden width & 64 \\
Activation & GELU \\
Optimizer & AdamW \\
Learning rate & $3\times10^{-4}$ \\
Weight decay & $10^{-3}$ \\
Loss & SmoothL1 \\
SmoothL1 $\beta$ & 0.01 \\
\bottomrule
\end{tabular}
\end{table}
Hidden features are standardized using the mean and standard deviation of the fitting split, and position targets are scaled as $p/8191$.
The full-attention comparison in Fig.~\ref{fig:dense-probe}
uses the same probe architecture and fitting procedure.
Decodability summarizes the reduction over a constant-position predictor with MAE $B_\ell=2016$ tokens:
\begin{equation}
 D=\frac{100}{16}\sum_{\ell=0}^{15}
 \left(1-\frac{\mathrm{MAE}_\ell}{B_\ell}\right),
\end{equation}
where $\ell$ indexes decoder-block outputs (block 0 is the first decoder block, not the embedding layer); higher $D$ means more accurate position recovery.

\subsection{Frozen Position Probes beyond the Training Length}

For the 345M step-30k checkpoints of HiLS-NoPE, HiLS-T5, and HiLS-HoPE, we fit a separate two-layer MLP to the output of each decoder block. The probes use raw-token positions 128--8191 from 64 training sequences, sampled every 16 tokens. We freeze the probes and evaluate them every 64 tokens on 32K prefixes of 16 held-out sequences. Feature normalization and position scaling remain fixed after fitting within 8K, and predictions are not clipped to the training range. Figure~\ref{fig:frozen-position-heatmap-345m} shows absolute-position
MAE divided by 32k on a shared color scale.

\section{Attention Dilution across Context Lengths}
\label{app:entropy-protocol}
\subsection{Full-Context Entropy Computation}
\label{app:native-entropy-audit}

For Full Attention, we use the causal attention probabilities directly. For HiLS, normalized nonnegative mixture weights $\pi_{L,c}$ and token probabilities $p_L(t\mid c)$ give
\begin{equation}
a_{L,t}=\sum_c\pi_{L,c}p_L(t\mid c),
\end{equation}
where $c$ indexes the local branch or a selected chunk, and $p_L(t\mid c)=0$ outside its real-token support.

For NSA, we spread mean-pooled compressed-value weights over the underlying tokens, combine the three sigmoid-gated branches, and renormalize their real-token mass; DashAttention uses its native merged raw-token weights. 

Inputs are paired by content and by position relative to the end: each length-$L$ input is the last $L$ tokens of one of 100 passages, and its final 16 tokens are the queries. Longer inputs only add earlier text, so the queries and the final 8K tokens are identical at every length.

For each passage $i$, query $q$, layer $\ell$, and head $h$, we pair the length-$L$ entropy with the same passage's 8K reference:
\begin{equation}
 \begin{aligned}
 \Delta_H^{\mathrm{native}}(L)
 &=\frac1{100}\sum_{i=1}^{100} \\
 &\quad\mathbb E_{q,\ell,h}
 \left[
 \left|H_{i,q,\ell,h}(L)-H_{i,q,\ell,h}(8K)\right|
 \right].
 \end{aligned}
 \label{eq:native-entropy-drift}
\end{equation}
Absolute differences are taken before averaging over query tokens, heads, and decoder layers 0--15, so increases and decreases do not cancel. Entropy is measured in nats, with $0\log 0=0$.

Figure~\ref{fig:decodability-entropy} pairs the 64K entropy drift in Eq.~\eqref{eq:native-entropy-drift} with the step-30k within-window decodability defined in Appendix~\ref{app:position-decodability}.

\subsection{Logarithmic Query Scaling}
\label{app:concentration-adjustment}
We apply logarithmic query scaling to Full-HoPE at step 30k, and evaluate the same RULER prompts as Table~\ref{tab:ruler-long} following the evaluation protocol in Appendix~\ref{sec:eval}. We scale the normalized queries before rotary encoding using Eq.~\eqref{eq:log-query-scaling}. Scaling applies to all query heads in layers 0--15; queries within the first 8,192 positions are unchanged. The coefficient 0.1 is fixed; weights and HoPE frequencies are unchanged.

\begin{table}[t]
\centering
\small
\setlength{\tabcolsep}{3pt}
\caption{Answer-token EM (\%) of Full-HoPE with and without logarithmic query scaling. SN, MK, and VT denote Single-NIAH, Multi-key, and Variable Tracking, respectively.}
\label{tab:log-scaling-complete}
\begin{tabular}{@{}lrrrrrr@{}}
\toprule
 & \multicolumn{3}{c}{Unscaled} & \multicolumn{3}{c}{Log scaling} \\
\cmidrule(lr){2-4}\cmidrule(lr){5-7}
Length & SN & MK & VT & SN & MK & VT \\
\midrule
16K  & 99 & 92 & 17 & 100 & 98 & 19 \\
32K  & 75 &  4 &  5 & 100 & 97 & 18 \\
64K  &  1 &  0 &  0 & 100 & 83 & 21 \\
128K &  0 &  0 &  0 &  98 & 48 &  6 \\
\bottomrule
\end{tabular}
\end{table}

\section{Evidence Selection in Sparse Attention}
\label{app:depth-selection}

Each task uses 100 RULER prompts per length from 8K to 64K. Since NSA always selects the current block, we filter prompts to keep all required evidence outside the current block and evaluate NSA's content ranking over historical blocks. We retain the first 100 qualifying prompts in dataset order for each task and length and use the same prompts across models. Each prompt has one gold evidence span for SN, two for MK,
and three for VT.

For Single-NIAH and Multi-key, we measure selector scores at the query position used to predict each target's first answer token. For Variable Tracking, we use the final variable in a three-statement chain and its three required evidence spans.

\paragraph{Metric definitions.}

For target case $i$, let $\mathcal G_i$ denote its set of required evidence spans, and let $T^{(k)}_{i,\ell,g}$ denote the at most $k$ eligible blocks with the highest native retrieval scores at layer $\ell$ and routing group $g$, before mandatory block-priority overwrites. For an evidence span $e$, we define
\begin{equation}
 A_i^{(k)}(e)=\mathbf1\!\left[
 \exists\ell,g\;\exists b\in T^{(k)}_{i,\ell,g}:
 e\cap B_b\ne\varnothing
 \right],
\end{equation}
where $B_b$ denotes the raw-token positions covered by physical block $b$. Pooling routing groups and decoder layers 0--15 for each evidence span gives the micro-averaged recall over gold evidence spans within each model/task/length setting:
\begin{equation}
 \mathrm{Recall}_{\mathrm{any}}@k=
 \frac{\sum_i\sum_{e\in\mathcal G_i}A_i^{(k)}(e)}
      {\sum_i|\mathcal G_i|}.
 \label{eq:any-layer-recall}
\end{equation}
We report $\mathrm{Recall}_{\mathrm{any}}@1$ as R@1 and
$\mathrm{Recall}_{\mathrm{any}}@8$ as Coverage@8.

\paragraph{Native selector scores.}
Ranks use raw pre-priority scores under native causal eligibility. For DashAttention, blocks with zero selector scores are excluded
from ranking. Ties are broken by block ID.

Figure~\ref{fig:decodability-r1} pairs 64K R@1, averaged across the three tasks, with the step-30k within-window decodability defined in Appendix~\ref{app:position-decodability}.

\section{Related Work}
\label{app:related-work}

\paragraph{Positional generalization.}
\citet{kazemnejad2023position} compare positional encodings on reasoning and mathematical tasks and find that NoPE generalizes well to longer inputs in this setting. Using position probes, \citet{puvvada2025swan} find that absolute positions can be decoded from NoPE hidden states within the training window, but that these predictions become less accurate beyond it.

\paragraph{Attention entropy stability.}
\citet{wang2024length} link length-generalization failures in NoPE to attention dispersion. Log-length scaling~\cite{chiang2022limitation} and adaptive temperatures~\cite{velickovic2025softmax} can sharpen softmax attention as inputs grow. \citet{vasylenko2026long} study how restricting attention support with $\alpha$-entmax can limit this dispersion.

\paragraph{Evidence retrieval in sparse attention.}
HiLS~\cite{hu2026hils} demonstrates strong length extrapolation with end-to-end learned chunk selection. \citet{leng2026length} identify an expressive chunk encoder, a bypassing residual path, and enforced selection sparsity during pretraining as key design principles for length generalization in hierarchical sparse attention.

Our study connects these lines of work through two controlled comparisons. Across three HiLS configurations, we relate absolute-position decodability within 8K to both entropy drift and evidence ranking at 64K. With HoPE fixed, we compare entropy drift across attention architectures and examine how highly each sparse selector ranks the required evidence as context grows.
\end{document}

%% file: figures/overview.tex
\begin{tikzpicture}[
  font=\sffamily\footnotesize,
  body/.style={
    draw=#1!65!black,
    line width=0.6pt,
    fill=#1!7,
    rounded corners=2.5pt,
    inner sep=5pt,
    anchor=north west
  },
  head/.style={
    fill=#1!65!black,
    text=white,
    rounded corners=2.5pt,
    text width=4.7cm,
    inner sep=5pt,
    anchor=south west,
    font=\sffamily\footnotesize\bfseries,
    align=left
  },
  arr/.style={
    -{Latex[length=2mm]},
    line width=0.7pt,
    draw=black!50
  },
]
\definecolor{cpos}{RGB}{196,110,30}
\definecolor{cent}{RGB}{40,104,158}
\definecolor{csel}{RGB}{50,125,82}
\definecolor{okc}{RGB}{35,125,65}
\definecolor{noc}{RGB}{185,45,40}

\newcommand{\bx}[1]{%
  \parbox{4.7cm}{\centering\strut #1\strut}%
}

\node[body=cpos] (b1) at (0,0)
  {\bx{\textit{Positional behavior generalizes}}};
\node[head=cpos] at (b1.north west)
  {1\enspace Positional generalization};

\node[body=cent] (b2) at (5.45,0)
  {\bx{\textit{Attention entropy stays stable}}};
\node[head=cent] at (b2.north west)
  {2\enspace Attention entropy stability};

\node[body=csel] (b3) at (10.9,0)
  {\bx{\textit{Evidence retains high priority}}};
\node[head=csel] at (b3.north west)
  {3\enspace Evidence retrieval accuracy};

\draw[line width=0.7pt, draw=black!50]
  ($(b1.north)+(0,0.62)$)
  -- ++(0,0.3)
  -| ($(b3.north)+(0,0.92)$);

\draw[arr]
  ($(b2.north)+(0,0.92)$)
  -- ($(b2.north)+(0,0.62)$);

\draw[arr]
  ($(b3.north)+(0,0.92)$)
  -- ($(b3.north)+(0,0.62)$);

\node[
  fill=white,
  inner sep=2pt,
  font=\sffamily\scriptsize,
  text=black!70
] at ($(b1.north)!0.5!(b2.north)+(0,0.92)$)
  {shapes};

\begin{scope}[shift={(2.6,-1.3)}]
  \def\ok{\textcolor{okc}{\normalsize$\checkmark$}}
  \def\no{\textcolor{noc}{\normalsize$\times$}}
  \def\na{\textcolor{black!45}{--}}
  \def\pa{\textcolor{cpos!85!black}{\normalsize$\sim$}}
  \def\rh{0.42}

  \fill[csel!12]
    (-0.15,{-7*\rh-0.2})
    rectangle (11.2,{-7*\rh+0.2});

  \node[anchor=base west,
        font=\sffamily\footnotesize\bfseries]
    at (0,-0.07) {Model};

  \node[anchor=base,
        font=\sffamily\footnotesize\bfseries,
        text=cpos!65!black]
    at (4.2,-0.07) {1 Position};

  \node[anchor=base,
        font=\sffamily\footnotesize\bfseries,
        text=cent!65!black]
    at (6.0,-0.07) {2 Entropy};

  \node[anchor=base,
        font=\sffamily\footnotesize\bfseries,
        text=csel!65!black]
    at (7.8,-0.07) {3 Evidence};

  \node[anchor=base,
        font=\sffamily\footnotesize\bfseries]
    at (10.0,-0.07) {Extrapolates};

  \draw[black!45] (-0.15,-0.22) -- (11.2,-0.22);

  \foreach \m/\a/\b/\cc/\r [count=\j] in {
    Full attention(HoPE)/\ok/\no/\na/\no,
    DashAttention(HoPE)/\ok/\no/\no/\no,
    NSA(HoPE)/\ok/\ok/\no/\no,
    HiLS(RoPE)/\no/\no/\no/\no,
    HiLS(NoPE)/\no/\no/\no/\no,
    HiLS(T5)/\pa/\pa/\pa/\pa,
    HiLS(HoPE)/\ok/\ok/\ok/\ok}
  {
    \node[anchor=west] at (0,{-\j*\rh}) {\m};
    \node at (4.2,{-\j*\rh}) {\a};
    \node at (6.0,{-\j*\rh}) {\b};
    \node at (7.8,{-\j*\rh}) {\cc};
    \node at (10.0,{-\j*\rh}) {\r};
  }

  \draw[black!45]
    (-0.15,{-7*\rh-0.22})
    -- (11.2,{-7*\rh-0.22});
\end{scope}
\end{tikzpicture}